\documentclass[sigconf]{acmart}
\usepackage{multirow}
\usepackage{threeparttable}
\usepackage{overpic}
\usepackage{algorithm,algorithmic}

\usepackage{graphicx,subfigure}

\def\E{\mathbf{E}}

\def\x{\mathbf{x}}

\def\y{\mathbf{y}}

\def\ttheta{\mbox{\boldmath$\theta$}}

\newcommand\blfootnote[1]{%
  \begingroup
  \renewcommand\thefootnote{}\footnote{#1}%
  \addtocounter{footnote}{-1}%
  \endgroup
}

\usepackage{enumitem}
\setlist{topsep=0pt, leftmargin=*}
\AtBeginDocument{%
  \providecommand\BibTeX{{%
    \normalfont B\kern-0.5em{\scshape i\kern-0.25em b}\kern-0.8em\TeX}}}

\setcopyright{acmcopyright}
\copyrightyear{2018}
\acmYear{2018}
\acmDOI{10.1145/1122445.1122456}

\acmConference[Woodstock '18]{Woodstock '18: ACM Symposium on Neural
  Gaze Detection}{June 03--05, 2018}{Woodstock, NY}
\acmBooktitle{Woodstock '18: ACM Symposium on Neural Gaze Detection,
  June 03--05, 2018, Woodstock, NY}
\acmPrice{15.00}
\acmISBN{978-1-4503-XXXX-X/18/06}

\begin{document}

\title{Multi-Epoch learning with Data Augmentation for Deep Click-Through Rate Prediction}


\author{Zhongxiang Fan$^{1*}$, Zhaocheng Liu$^{1*}$, Jian Liang$^{1*}$, Dongying Kong$^1$, Han Li$^1$, Peng Jiang$^1$, Shuang Li$^2$, Kun Gai$^3$}
\affiliation{%
  \institution{$^1$Kuaishou Technology \quad $^2$Beijing Institute of Technology \quad $^3$Unaffiliated}
  \and 
  {fanzhongxiang, liuzhaocheng, liangjian03, kongdongying, lihan08, jiangpeng}@kuaishou.com
  \and
  shuangli@bit.edu.cn \quad  gai.kun@qq.com
  \country{}
}

\renewcommand{\authors}{Zhongxiang Fan, Zhaocheng Liu, Jian Liang, Dongying Kong, Han Li, Peng Jiang, Shuang Li, Kun Gai}

\renewcommand{\shortauthors}{Fan, Liu, and Liang, et al.}

\begin{abstract}
This paper investigates the one-epoch overfitting phenomenon in Click-Through Rate (CTR) models, where performance notably declines at the start of the second epoch. Despite extensive research, the efficacy of multi-epoch training over the conventional one-epoch approach remains unclear. We identify the overfitting of the embedding layer, caused by high-dimensional data sparsity, as the primary issue. To address this,
we introduce a novel and simple Multi-Epoch learning with Data Augmentation (MEDA) framework, suitable for both non-continual and continual learning scenarios, which can be seamlessly integrated into existing deep CTR models and may have potential applications to handle the ``forgetting or overfitting'' dilemma in the retraining and the well-known catastrophic forgetting problems.
MEDA minimizes overfitting by reducing the dependency of the embedding layer on subsequent training data or the Multi-Layer Perceptron (MLP) layers, and achieves data augmentation through training the MLP with varied embedding spaces.
Our findings confirm that pre-trained MLP layers can adapt to new embedding spaces, enhancing performance without overfitting. This adaptability underscores the MLP layers' role in learning a matching function focused on the relative relationships among embeddings rather than their absolute positions.
To our knowledge, MEDA represents the first multi-epoch training strategy tailored for deep CTR prediction models.
We conduct extensive experiments on several public and business datasets, and the effectiveness of data augmentation and superiority over conventional single-epoch training are fully demonstrated.
Besides, MEDA has exhibited significant benefits in a real-world online advertising system. \blfootnote{$*$ Equal contributions from the three authors.}
\end{abstract}

\ccsdesc[500]{Information systems~Recommender systems}

\keywords{Click-Through Rate Prediction, Overfitting, Multi-Epoch Learning}


\maketitle

\section{Introduction}
\label{section:intro}
Click-through rate (CTR) prediction is crucial in online recommendation and advertising systems, benefiting significantly from advancements in deep learning-based models~\cite{cheng2016wide,qu2016product,guo2017deepfm,yu2020deep,zhou2018deep,zhou2019deep,pi2019practice,li2022adversarial}. Despite the progress and diverse approaches, including non-continual learning for smaller datasets and continual learning~\cite{cai2022reloop,guan2022deployable,mi2020ader,yang2023incremental} for larger or real-time datasets, a common challenge persists: ``one-epoch overfitting''~\cite{zhou2018deep,zhang2022towards}.
This phenomenon, where model performance drops sharply at the beginning of the second training epoch, contrasts with other deep learning domains like computer visionn~\cite{he2016deep,russakovsky2015imagenet}, audio processing~\cite{purwins2019deep}, and natural language processing~\cite{devlin2018bert,vaswani2017attention}, where multiple epochs enhance model convergence.
Specifically, for industrial areas with sufficient computation resources and confronted with model-convergence issues (e.g., with small datasets), multi-epoch learning is valuable. Furthermore, multi-epoch learning allows necessary ``rethinking'', exemplified in tasks like unsupervised domain adaptation~\cite{wilson2020survey} or label-noise correction~\cite{song2022learning}. Therefore, addressing the one-epoch overfitting issue is crucial for improving the efficacy of CTR prediction models in industrial settings.

Unfortunately, most research~\cite{belkin2019reconciling,salman2019overfitting,zhou2021over,zhang2021understanding,arpit2017closer} on the overfitting problem of Deep Neural Networks (DNNs) has predominantly focused areas outside CTR prediction.
Moreover, to our knowledge, in addition to the CTR domain, only large language models undergoing supervised fine-tuning have reported similar one-epoch overfitting phenomena~\cite{ouyang2022training,komatsuzaki2019one}.
Despite recent efforts~\cite{zhang2022towards} to dissect the underlying factors of this phenomenon within deep CTR models, the potential benefits of adopting a multi-epoch training paradigm remain unexplored, as optimal performance is often attained with a single epoch of training.

\begin{figure}
    \centering
    \includegraphics[width=\linewidth]{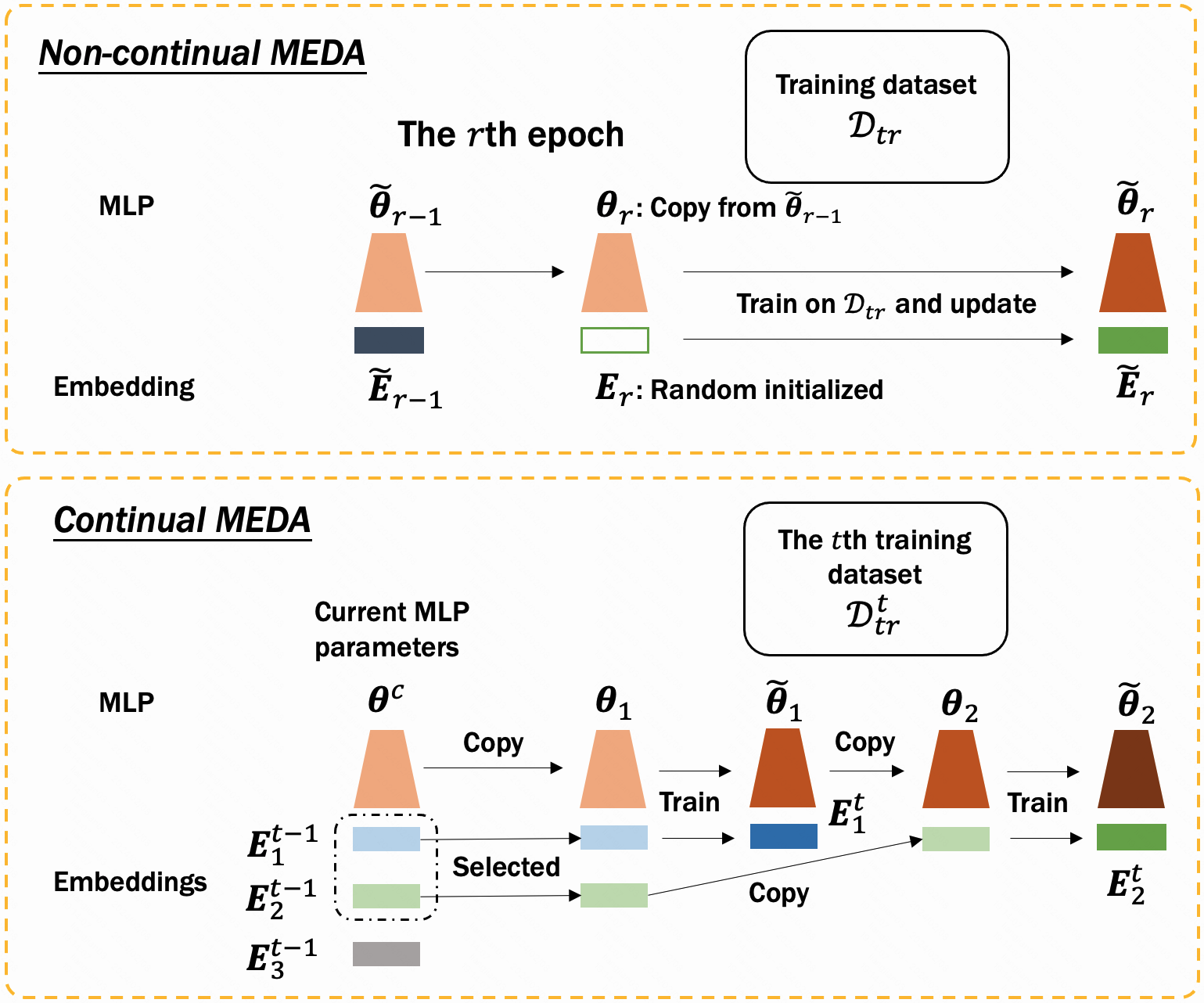}
    \caption{Our proposed MEDA framework. For non-continual learning, MEDA reinitializes the embedding parameters at the onset of each training epoch; for continual learning, MEDA maintains multiple independently initialized embedding layers and for each dataset, trains each embedding layer once successively. The embedding layers can be selected based on requirements or costs.
    }
    \label{fig:multi_epoch_learning}
\end{figure}
 
To address the one-epoch overfitting challenge in CTR models, we introduce a novel and simple Multi-Epoch learning with Data Augmentation (\textbf{MEDA}) framework, tailored for both non-continual and continual learning scenarios. Our framework can also cover both the classification and regression tasks. Specifically, we identify the overfitting of the embedding layer, caused by high-dimensional data sparsity, as the primary issue for one-epoch overfitting, then design MEDA to effectively mitigate overfitting by decoupling the embedding and Multi-Layer Perceptron (MLP) layers in CTR models~\cite{zhang2016deep}, addressing the core issues of high data dependency and low occurrence rates in embedding vectors that lead to poor generalization on test datasets, and reducing the embedding-data and embedding-MLP dependencies. In detail, in our non-continual MEDA algorithm shown in Figure \ref{fig:multi_epoch_learning}, the embedding-data dependency is reduced by reinitializing the embedding layer at the onset of each training epoch. The non-continual MEDA is extended to the continual MEDA to further reduce the additional embedding-MLP dependency in the continual learning setting, which is shown in Figure \ref{fig:multi_epoch_learning}. We leverage multiple independently initialized embedding layers---each for an extra epoch: for each dataset, each embedding layer can be selected to train once successively. The selection can be based on business requirements or training costs. Various variants of our continual MEDA can be designed for specific requirements. Intuitively, on each dataset, each embedding layer in MEDA is trained once only, thereby minimizing overfitting, while the MLP layers are trained repeatedly to improve convergence. Our proposed MEDA can be regarded as a data augmentation method because it can be treated as learning the same categorical features with different embedding spaces. To our knowledge, MEDA represents the first multi-epoch training strategy tailored for deep CTR prediction models. Potential applications of our MEDA may be designed to handle the ``forgetting or overfitting'' dilemma in the \emph{retraining}~\cite{wang2023gradient} and the well-known \emph{catastrophic forgetting}~\cite{katsileros2022incremental,hu2022continual} problems, in addition to the normal multi-epoch training task.

We conduct comprehensive experiments on public and business datasets to show the effectiveness of data augmentation and superiority over single-epoch learning. 
Notably, MEDA's second-epoch performance consistently exceeds that of single-epoch training across various datasets and CTR models, with improvements in test AUC ranging from 0.8\% to 4.6\%. This trend persists across multiple epochs without inducing overfitting, offering flexibility in training duration based on cost considerations. Our findings confirm that pre-trained MLP layers can adapt to new embedding spaces, enhancing performance without overfitting. This adaptability underscores the MLP layers' role in learning a matching function focused on the relative relationships among embeddings rather than their absolute positions. Furthermore, MEDA demonstrates remarkable efficiency by achieving or surpassing the outcomes of complete data single-epoch training with only a fraction of the data, e.g., in most cases, MEDA with $1/2$ data can outperform single-epoch training with complete data, sometimes even $3$ epochs on $1/8$ data can outperform $1$ epoch on complete data, indicating its potential for significant data augmentation benefits. 
The successful deployment of MEDA in a live environment, corroborated by positive online A/B testing results, further attests to its practical value and impact.

\section{Related Works}
 
\noindent\textbf{\emph{\underline{One-Epoch Overfitting}}}.
Recent empirical research~\cite{zhang2022towards} indicates that reducing feature sparsity can diminish the prevalence of one-epoch overfitting, yet the potential superiority of multi-epoch training over traditional single-epoch approaches remains uncertain. Additionally, established anti-overfitting strategies such as weight decay and dropout have proven ineffective in this context~\cite{zhang2022towards}. Parallel observations~\cite{ouyang2022training} in large language models undergoing supervised fine-tuning reveal a similar tendency towards one-epoch overfitting, albeit with a suggestion that a moderate level of overfitting might actually benefit downstream tasks. This concept of "appropriate overfitting" presents an intriguing avenue for future exploration within our proposed framework.

\noindent\textbf{\emph{\underline{Pre-training}}}.
Recent approaches~\cite{lin2023map,liu2022boosting,wang2023bert4ctr,muhamed2021ctr} have explored pre-training to enhance the representational capabilities of embedding and feature extraction layers within MLPs for various applications, yet these advancements fall short in demonstrating their efficacy in avoiding overfitting when CTR prediction is incorporated as an auxiliary training objective, nor do they facilitate multi-epoch training for such models. In contrast, graph learning research has delved into fine-tuning pre-trained models for new graphs, facing challenges related to either maintaining a consistent node ID space~\cite{hu2019strategies,liu2023graphprompt,lu2021learning,hu2020gpt} or solely leveraging graph structure while neglecting node features~\cite{qiu2020gcc,zhu2021transfer}. This leaves an open question in the context of CTR models: the potential for pre-trained MLP layers to contribute positively to a distinct embedding space remains unexplored and warrants further investigation.

\section{Background}

\noindent\textbf{\emph{\underline{Features}}}.
CTR (Click-Through Rate) prediction models are distinguished by their handling of high-dimensional sparse data, often involving billions of categorical features with low occurrence rates~\cite{jiang2019xdl,zhao2019aibox,zhao2020distributed}. These features generally encompass~\cite{cheng2016wide,qu2016product,guo2017deepfm,yu2020deep,zhou2018deep,zhou2019deep,pi2019practice,li2022adversarial} item profiles (including Item ID, brand ID, shop ID, category ID, etc.), user profiles (User ID, age, gender, income level), and both long- and short-term user behaviors, reflecting a mix of enduring interests and immediate needs. Prior to modeling, raw features undergo discretization~\cite{liu2020empirical} to transform numerical inputs into categorical ones and feature selection~\cite{guo2022lpfs} to refine the dataset, resulting in predominantly categorical inputs. In real-world industrial applications, categorical features are typically extremely high-dimensional sparse~\cite{jiang2019xdl,zhao2019aibox,zhao2020distributed}.

\noindent\textbf{\emph{\underline{Embedding}}}.
To handle these categorical features, most deep CTR prediction models share a similar Embedding and Multi-Layer Perceptron (MLP) architecture.
Specifically, they typically adopt an embedding layer~\cite{zhang2016deep} at the front, followed by various types of MLP structures, with the embedding layer responsible for mapping the high-dimensional categorical features to low-dimensional vectors.

\noindent\textbf{\emph{\underline{MLP}}}.
Given the concatenated dense representation vector, an MLP is employed to capture the nonlinear interaction among features~\cite{liu2020dnn2lr}.
Numerous studies~\cite{cheng2016wide,qu2016product,guo2017deepfm,yu2020deep} have concentrated on devising MLP architectures that excel in extracting pertinent information for tabular data.
It is worth noting that, given the rapid expansion of user historical behavior data in real-world industrial applications, careful attention must be paid to user behavior modeling during the design of MLPs.
The relevant methods~\cite{zhou2018deep,zhou2019deep,pi2019practice,li2022adversarial} considering user behavior modeling focus on capturing the dynamic nature of user interests based on their historical behaviors which typically consist of lists of categorical features.
 
\section{Methodology}
\label{sec:method}

In this section, we present the detailed methodology of our method. First, we define the notations and problem settings of our study.

\noindent\textbf{\emph{\underline{Non-continual Learning}}}.
Consider a dataset $\mathcal{D} = \{(\x^i,\y_i)\}_{i=1}^n$ consisting of $n$ independent samples. For the $i$th sample, $\x^i\in\mathcal{X}\subset\mathbb{R}^d$ is a feature vector with $d$ dimensions, $y^i\in\mathcal{Y}\subset\mathbb{R}^m$ is the label of the $i$th sample. Note that our framework can cover both classification and regression tasks. Let $M\in\mathcal{M}$ be a data-driven model. Specifically, let $\ttheta$ be the collection of training parameters of the MLP layers, and $\E$ be the collection of training parameters of the embedding layer.  Let $A\in\mathcal{A}$ be a training algorithm, and we denote by $M=A(\{\mathcal{D}:k\})$ as obtaining the model $M$ by training the dataset $\mathcal{D}$ by algorithm $A$ for $k\in \mathbb{Z}_+$ epochs. And we denote by $S(\mathcal{D}\mid M)\in \mathbb{R}$ as the evaluation score (the bigger the better) obtained from evaluating $M$ on $\mathcal{D}$. Finally, splitting $\mathcal{D}$ into $\mathcal{D}_{tr},\mathcal{D}_{te}$ as the training and testing datasets, respectively, our goal is to see if there exists a $k>1$ such that $S(\mathcal{D}_{te}\mid A(\{\mathcal{D}_{tr}:k\}))>S(\mathcal{D}_{te}\mid A(\{\mathcal{D}_{tr}:1\}))$.

\noindent\textbf{\emph{\underline{Continual Learning}}}.
Following the notations in the non-continual learning setting, consider the training dataset $\mathcal{D}_{tr} = \{\mathcal{D}_{tr}^t\}_{t=1}^T$ consisting of $T$ successive sub-datasets. Our goal is to see if there exists a $k^t>1$ such that $S(\mathcal{D}_{te}\mid A(\{\mathcal{D}_{tr}^t:k^t\}_{t=1}^T))>S(\mathcal{D}_{te}\mid A(\{\mathcal{D}_{tr}^t:1\}_{t=1}^T))$, where we denote by $A(\{\mathcal{D}_{tr}^t:k^t\}_{t=1}^T)$ as training each $\mathcal{D}_{tr}^t$ for $k^t$ epochs and denote by $A(\{\mathcal{D}_{tr}^t:1\}_{t=1}^T)$ as training each $\mathcal{D}_{tr}^t$ for $1$ epoch. The order of sub-datasets for training is not constrained. 

In the following, we first introduce the modeling of and {rationale} for our MEDA framework, then detail two instantiations to cover the non-continual and continual learning settings, respectively.

 
\subsection{Our MEDA Framework}

As introduced, for multi-epoch training, we propose MEDA to simultaneously maintain feature sparsity and avoid the one-epoch overfitting.  

Overfitting typically arises from a strong dependency between the training data and model parameters that fails to generalize to unseen data. Addressing overfitting involves either reducing this dependency or enhancing model generalization. In the context of CTR models, by decoupling the embedding from the MLP layers, we identify two main types of dependencies: embedding-data and embedding-MLP dependencies, where the latter views the low-dimensional embeddings as input data for the MLP layers. Our detailed experiments in Section~\ref{sec:exp} demonstrate that overfitting predominantly stems from the embedding-data interaction. Specifically, repeated training of the embedding layer on the same dataset leads to significant overfitting. In contrast, multiple training iterations over the same embedding vectors for the MLP layers result in much less or no overfitting.
This discrepancy is likely due to the sparse nature of high-dimensional data, where a vast number of categorical values exist but each appears infrequently. Consequently, embedding vectors, representing these infrequent values, are prone to overfitting due to limited training samples. Meanwhile, MLP parameters, engaging with the entire dataset, exhibit a lower risk of overfitting.

\noindent\textbf{\emph{\underline{Non-continual Problem Formulation}}}.
Building on the findings discussed, we approach the problem with the hypothesis that overfitting occurs when the initial parameters of the model align too closely with the exact information present in the training data. This insight leads us to the conclusion that optimizing initial parameter settings is crucial for mitigating overfitting. Notably, the phenomenon of one-epoch overfitting, which emerges at the beginning of the second training epoch, underscores that overfitting is triggered when the initial parameters precisely mirror the information within a specific data sample. To circumvent this, the initial parameters must be devoid of any exact information from future training samples. A straightforward solution is to randomize the initial parameters, ensuring their independence from subsequent training data. Given our analysis pinpointing embedding overfitting as the primary concern, we propose the novel strategy of randomly initializing embedding parameters at the start of each epoch in our non-continual MEDA framework. See Algorithm~\ref{alg_training} for details. 

Conventional wisdom might suggest that reinitializing embedding parameters could disrupt the MLP's ability to recognize embeddings, potentially undermining training and prediction efficacy. However, our experimental findings detailed in Section~\ref{sec:exp} reveal a different outcome: the MLP not only successfully identified the embeddings but also yielded superior prediction accuracy. This indicates that the MLP was able to discern the relative relationships among embeddings. Despite significant differences in the final embedding parameters across epochs, as shown in Section~\ref{sec:exp}, the essential insight is that for CTR model MLP layers, the precise values or absolute positions of embeddings are less critical than their interrelations. This understanding allows us to view the additional embeddings processed by the MLP as augmented data samples. These samples maintain crucial semantic relationships within varying embedding spaces, effectively serving as a form of data augmentation. Furthermore, our results demonstrate that the MLP was nearing convergence throughout the multi-epoch training, suggesting that despite initial embedding values differing between epochs, the semantic essence crucial for CTR prediction was consistently captured in our multi-epoch learning framework.

\begin{algorithm}[t!]
    \caption{Non-continual MEDA}
    \label{alg_training}
    \begin{algorithmic}[1]
        \REQUIRE Training dataset $\mathcal{D}_{tr}$, training algorithm $A$,  the number of training epoch $k$.
        \ENSURE MLP parameters $\hat{\ttheta}$ and embedding parameters $\hat{\E}$.
        \STATE Random initialize $\tilde{\ttheta}_0$.
        \FOR{epoch $r=1$  to  $k$}
        \STATE Initialization: Random initialize $\E_r$. $\ttheta_{r}=\tilde{\ttheta}_{r-1}$.
        \STATE Training and Update: $\tilde{\ttheta}_r,\tilde{\E}_r=A(\{\mathcal{D}_{tr}:1\})$ with $\ttheta_r,\E_r$ as the initial parameters.
        \ENDFOR
        \RETURN $\hat{\ttheta} = \tilde{\ttheta}_{k}$, $\hat{\E}=\tilde{\E}_k$.
    \end{algorithmic}
\end{algorithm}

\noindent\textbf{\emph{\underline{Continual Problem Formulation}}}.
In a continual learning framework, where datasets are processed successively, we encounter a unique challenge: one-epoch overfitting also occurs upon the second training of the $t$th dataset $t>1$, involving both embedding and MLP layer optimization. This scenario diverges from the non-continual setting, as reinitializing embedding parameters at the start of the $t$th dataset's training effectively disregards the accumulated knowledge from datasets $1$ to $(t-1)$, which is undesirable. Drawing from insights in the non-continual setting, to prevent one-epoch overfitting,
the initial embedding parameters should not contain \emph{exact} information in any data sample in the $t$th dataset, but should contain information in datasets $1\sim (t-1)$. 
Therefore, one option involves adopting the final embedding parameters from the $t-1$th dataset's training $\E^{t-1}$. Nonetheless, given that the current MLP layers $\ttheta^c$ have already interacted with $\E^{t-1}$ during the $t$th dataset training, a high dependency exists between $\ttheta^c$ and $\E^{t-1}$, heightening the risk of overfitting, as our experiments in Section~\ref{sec:exp} corroborate, even if we fix $\E^{t-1}$ for training.
 
Therefore, we further propose to reduce the dependency between $\ttheta^c$ and the initial embedding parameters for the second-epoch training of dataset $t$, which belongs to the second type of data-parameter dependency. Specifically, a group of embedding parameters with a different embedding space and trained on datasets $1\sim (t-1)$ may satisfy the constraints above. Thus we combine the solution in non-continual MEDA to independently initialize multiple groups of embedding parameters to form distinct embedding spaces. These are then trained sequentially with the MLP on each dataset. See Algorithm~\ref{alg_training_continual} for details. 
Note that our non-continual and continual MEDA algorithms are consistent: our non-continual can be regarded as a special case of our continual MEDA for a singular dataset scenario.

Comparing our non-continual MEDA and continual MEDA for the same training and testing dataset, i.e., continual MEDA will split the training datasets, the non-continual MEDA exhibits intuitive superiority because the differences between different groups of embedding parameters are larger than those in the continual MEDA. 
Consider an extreme situation of continual MEDA such that each data sample is a dataset, since the values of initial embedding parameters are usually small, the gradients for different groups of embedding parameters may be similar, which will result in similar embedding parameters. This similarity could diminish the intended effect of data augmentation achieved by training the MLP with diverse embedding spaces, which is also supported by the results in Section~\ref{sec:exp}.

\begin{algorithm}[t!]
    \caption{Continual MEDA}
    \label{alg_training_continual}
    \begin{algorithmic}[1]
        \REQUIRE Training dataset $\mathcal{D}_{tr} = \{\mathcal{D}_{tr}^t\}_{t=1}^T$, training algorithm $A$,  the max number of training epoch $k$.
        \ENSURE MLP parameters $\hat{\ttheta}$ and embedding parameters $\hat{\E}$.
        \STATE Random initialize $\ttheta^{c},\{\E_r^0\}_{r=1}^k$.
        \FOR{ dataset index $t=1$  to  $T$}
        \FOR{epoch $r=1$  to  $k$}
        \IF{$\E_r^{t-1}$ is selected based on business requirements and training costs}
        
        \STATE Initialization: $\ttheta_{r}=\ttheta^{c}$.
       
        \STATE Training and Update: $\tilde{\ttheta}_r,\E_r^t=A(\{\mathcal{D}_{tr}^t:1\})$ with $\ttheta_r,\E_r^{t-1}$ as the initial parameters.
        \STATE  $\ttheta^c = \tilde{\ttheta}_r$, $\E^c=\E_r^t$.
        \ELSE 
        \STATE $ \E_r^t=\E_r^{t-1}$.
        \ENDIF
        \ENDFOR
        \ENDFOR
        \RETURN $\hat{\ttheta} = \ttheta^c$, $\hat{\E}=\E^c$.
    \end{algorithmic}
\end{algorithm}

\subsection{Potential Applications}
 
\noindent\textbf{\emph{\underline{Retraining}}}.
Retraining requirements~\cite{wang2023gradient} arise when there are significant alterations in data structure, features, or models, necessitating model updates to align with the revised pipeline. This scenario presents a "forgetting or overfitting" dilemma regarding the transfer of optimized model parameters from the old to the new model. Forgetting valuable information is a risk if parameters are discarded, whereas transferring them for retraining could lead to overfitting. Our non-continual MEDA approach addresses this challenge by transferring MLP for retraining to mitigate overfitting.

\noindent\textbf{\emph{\underline{Catastrophic Forgetting}}}.
The well-known issue of {catastrophic forgetting}~\cite{katsileros2022incremental,hu2022continual} in CTR prediction represents a decline in a model's ability to accurately predict outcomes for specific IDs not recently included in training. This phenomenon poses a significant challenge for maintaining accurate predictions on historical data. The ``forgetting or overfitting'' dilemma in this problem involves deciding whether to retrain the model using data from certain important IDs. The overfitting will also occur when retraining the data. Therefore, our continual MEDA may be modified for such retraining and then avoid overfitting.

\subsection{Computation/Storage Complexity Analyses}
Since our method only adds initialization processes, which are negligible for the computation complexity of training. Therefore, our method adds negligible computation complexity compared with standard multi-epoch learning. On the other hand, for the continual learning setting, our method requires $\mathcal{O}(kND)$ storage resources to maintain $k$ groups of embedding parameters for $k$-epoch learning, with $N$ representing the number of IDs and $D$ embedding vector dimension.

\section{EXPERIMENTS}\label{sec:exp}

In this section, we present the experimental setup and conduct extensive experiments to evaluate the effectiveness and superiority of our proposed MEDA framework, along with ablation studies and online A/B test results. 
\subsection{Experimental Setup}
\label{section:setup}

\begin{table}[t!]
\centering
\caption{Statistics of four datasets.}
\label{tab:datasets}
\scalebox{0.86}{
\begin{tabular}{c|c|c|c|c}
\hline
           & \#Records & \#users & \#items & \#categories   \\
\hline
    Amazon & 51M & 1.5M & 2.9M & 1252 \\  \hline
  Taobao & 89M & 1M & 4M & 9407  \\  \hline
  SVO & 1.75B & 0.2B & 6M & 1259  \\ \hline
  SVSL  & 0.3B & 40M & 0.5M & 324 \\  \hline
\end{tabular}
}
 
\end{table}

\noindent\textbf{\emph{\underline{Datasets}}}. 
We conduct comprehensive evaluations on two public datasets and two business datasets, with statistics in Table~\ref{tab:datasets}.
\par\textbf{Amazon dataset\footnote{\url{https://nijianmo.github.io/amazon/index.html}}}. It is a frequently used \emph{public} benchmark that consists of product reviews and metadata collected from Amazon~\cite{ni2019justifying}. In our study, we adopt the Books category of the Amazon dataset. 
We predict whether a user will review an item. 
\par\textbf{Taobao dataset\footnote{\url{https://tianchi.aliyun.com/dataset/649}}}. It is a \emph{public} compilation of user behaviors for CTR prediction from Taobao’s recommender system~\cite{zhu2018learning}. 
\par\textbf{Short-Video Order (SVO) dataset}. It is our collected \emph{large business} dataset that consists of user behaviors from Kuaishou’s recommender system. 
For this dataset, we predict the order behaviors of each user. We split 6 days for training and 1 day for testing. 
\par\textbf{Short-Video Search LTV (SVSL) dataset}. It is also our collected \emph{business} dataset that consists of user behaviors from Kuaishou’s \emph{search} system. 
For this dataset, we predict the Life-Time Value (LTV)~\cite{theocharous2015ad} value of each order for each user. We split 370 days for training and 1 day for testing.

For continual learning, for both the public and the SVO datasets, we split the first half of training data as $\mathcal{D}_{tr}^1$ and the rest as $\mathcal{D}_{tr}^2$, while for the SVSL dataset, we split the first 280 days of training data as $\mathcal{D}_{tr}^1$ and the rest as $\mathcal{D}_{tr}^2$. 

\noindent\textbf{\emph{\underline{CTR Models and Metrics for Evaluation}}}. 
We apply our method on the following CTR Models:
\begin{itemize}
\item \textbf{DNN} is a base deep CTR model, consisting of an embedding layer and a feed-forward network with ReLU activation.
\item \textbf{DIN}~\cite{zhou2018deep} proposes an attention mechanism to represent the user interests w.r.t. candidates.
\item \textbf{DIEN}~\cite{zhou2019deep} uses GRU to model user interest evolution.
\item \textbf{MIMN}~\cite{pi2019practice} proposes a memory network-based model to capture multiple channels of user interest drifting for long-term user behavior modeling.
\item \textbf{ADFM}~\cite{li2022adversarial} proposes an adversarial filtering model on long-term user behavior sequences.
\end{itemize}
For the business datasets, we adopt DIN as default. We denote our non-continual and continual MEDA as \textbf{MEDA-NC} and \textbf{MEDA-C}, respectively. Our methods equal single-epoch learning when the number of epoch is $1$.
For binary classification tasks, i.e., click or order prediction, we use Area under the curve (AUC) and binary cross-entropy loss as evaluation metrics, while for the regression tasks, i.e., the LTV prediction, we use AUC score between the LTV prediction scores and the binary order labels as the evaluation metric, following the common business practice.

\noindent\textbf{\emph{\underline{Implementation Details}}}. 
All CTR Models adhere to the optimal hyperparameters reported in their respective papers.
For public datasets, we adopt Adam~\cite{kingma2014adam} as the optimizer with a learning rate of 0.001, and Glorot~\cite{glorot2010understanding} as the initializer for embedding parameters.
For business datasets, we adopt Adagrad~\cite{duchi2011adaptive} as the optimizer with a learning rate of 0.01, and uniform initializer with the range of 0.01.


\subsection{Effectiveness and Superiority Evaluation}
\label{subsection:performance_eval}

\begin{figure}[t!]
\centering
\subfigure[Amazon]{\includegraphics[width=0.23\textwidth]{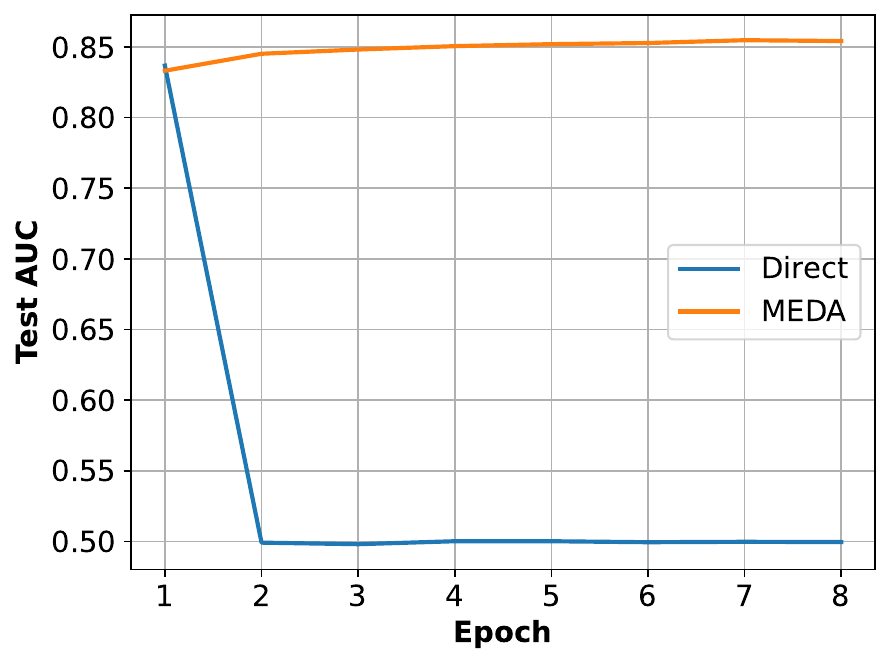}}
\subfigure[Taobao]{\includegraphics[width=0.23\textwidth]{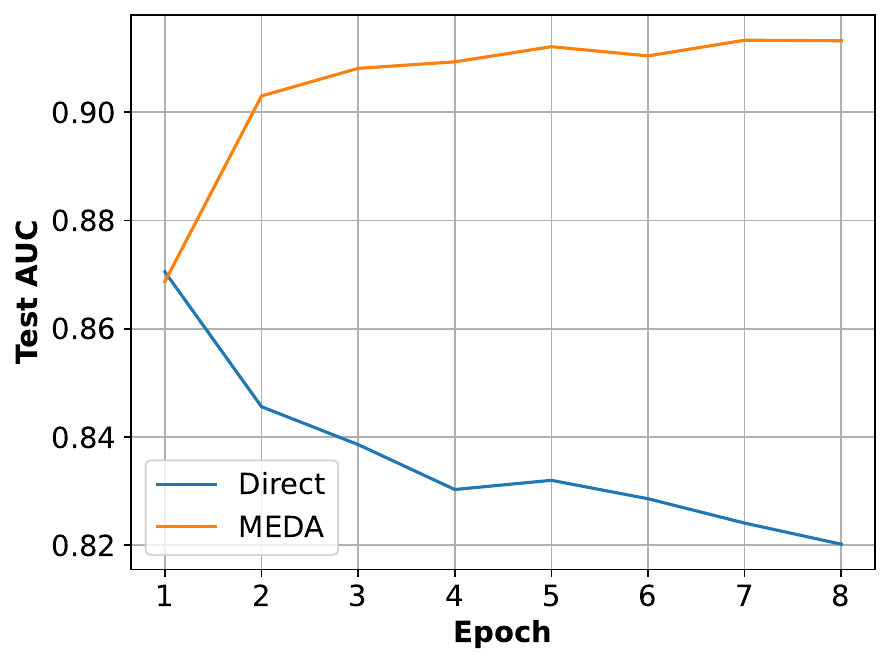}}
\caption{The test AUC curves of the Direct multi-epoch learning and our non-continual MEDA on the public datasets. }
\label{fig:show_problem}
\end{figure}

\noindent\textbf{\emph{\underline{Problem Justification}}}.
In Figure~\ref{fig:show_problem} highlights the presence and substantial impact of one-epoch overfitting. In both the Amazon and Taobao datasets, the test AUC rapidly declines starting from the second epoch of the direct multi-epoch learning. Whereas our MEDA can effectively improve the test AUC with the increase of epoch without overfitting. The overfitting issue is more pronounced in the Amazon dataset due to its higher data sparsity (less data, more IDs).

\begin{figure}[t!]
\centering
\subfigure[Amazon]{\includegraphics[width=0.23\textwidth]{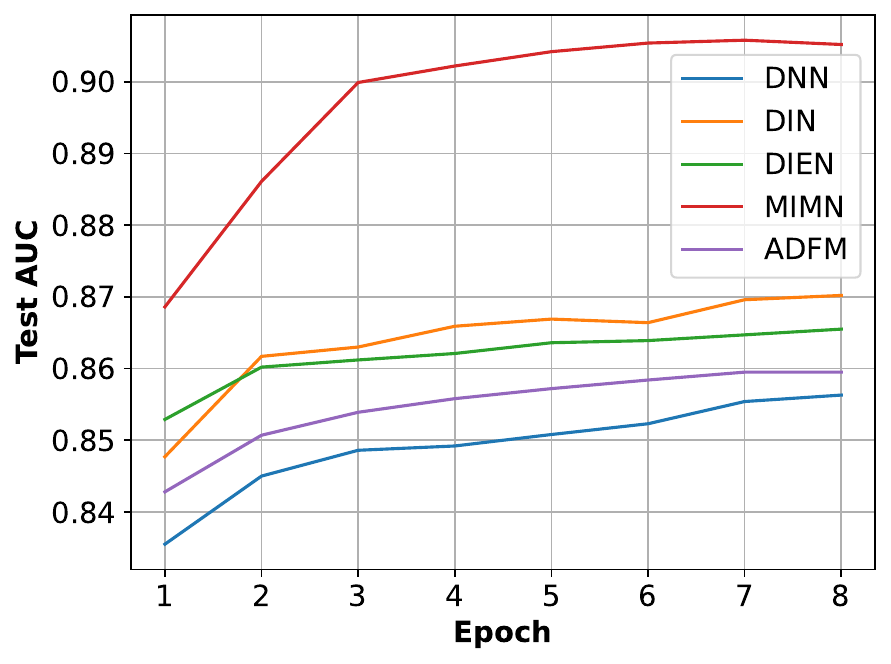}}
\subfigure[Taobao]{\includegraphics[width=0.23\textwidth]{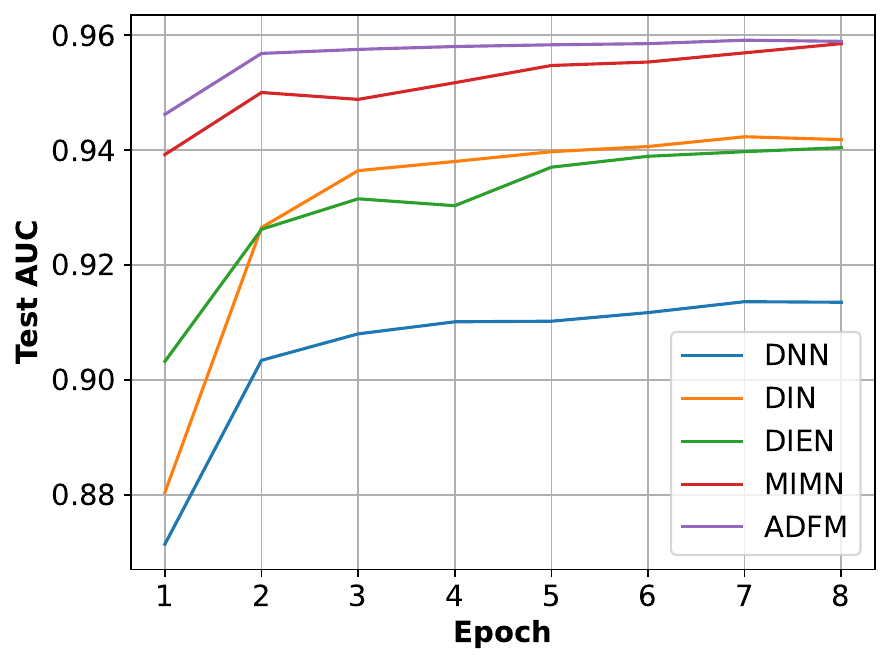}}
\caption{The test AUC curves of various models trained with our non-continual MEDA on the public datasets.}
\label{fig:ablation_study}
\end{figure}

\noindent\textbf{\emph{\underline{Superiority Evaluation}}}.
Tables~\ref{tab:second_epoch} and~\ref{tab:second_epoch_business} highlight the significant superiority of our MEDA approach over conventional single-epoch learning. Our non-continual and continual MEDA methods outperform single-epoch learning on both public and business datasets by a substantial margin, which aligns with the improvement magnitude of each CTR model. Furthermore, continual MEDA slightly outperforms non-continual MEDA on the Taobao dataset but slightly underperforms on the Amazon, SVO, and SVSL datasets. This suggests that data augmentation in continual MEDA is weaker, and data sparsity is more severe in these three datasets. The results presented for 
$2$ epochs of MEDA are reasonable for most industrial applications, considering computation and storage costs. Additionally, Figure \ref{fig:ablation_study} demonstrates stable increases in test AUCs for most models as the number of epochs increases. Hence, it is feasible to determine the stopping point at any epoch, as the AUC does not significantly decrease after a certain number of epochs. This user-friendly feature enables users to select the number of epochs based on training costs. Moreover, our MEDA approach outperforms single-epoch learning even more significantly when trained for 
$8$ epochs, further highlighting its superiority.

\begin{table}[t!]
\centering
\caption{The test AUC performance on the public datasets. MEDA methods run $2$ epochs.}
\label{tab:second_epoch}
\scalebox{0.86}{
\begin{tabular}{c|c|c|c|c|c|c}
\hline
           & & DNN & DIN & DIEN & MIMN & ADFM \\
\hline
\multirow{5}{*}{Amazon} & Single-Epoch & 0.8355 & 0.8477 & 0.8529 & 0.8686 & 0.8428 \\ \cline{2-7}
& MEDA-NC & \textbf{0.8450} & \textbf{0.8617} & \textbf{0.8602} & \textbf{0.8861} & \textbf{0.8507} \\ \cline{2-7}
& Improv. & +0.95\% & +1.4\% & +0.73\% & +1.75\% & +0.79\% \\ \cline{2-7}
& MEDA-C & \textbf{0.8446} & \textbf{0.8588} & \textbf{0.8587} & \textbf{0.8832} & \textbf{0.8516} \\ \cline{2-7}
& Improv. & +0.91\% & +1.11\% & +0.58\% & +1.46\% & +0.88\% \\ \hline
\multirow{5}{*}{Taobao} & Single-Epoch & 0.8714 & 0.8804 & 0.9032 & 0.9392 & 0.9462 \\ \cline{2-7}
& MEDA-NC & \textbf{0.9034} & \textbf{0.9265} & \textbf{0.9262} & \textbf{0.9500} & \textbf{0.9568} \\ \cline{2-7}
& Improv. & +3.2\% & +4.61\% & +2.3\% & +1.08\% & +1.06\% \\ \cline{2-7}
& MEDA-C & \textbf{0.9054} & \textbf{0.9321} & \textbf{0.9281} & \textbf{0.9565} & \textbf{0.9549} \\ \cline{2-7}
& Improv. & +3.40\% & +5.17\% & +2.49\% & +1.73\% & +0.87\% \\ \hline

\end{tabular}
}
\end{table}

\begin{table}[t!]
\centering
\caption{The test AUC performance on the business datasets. MEDA methods run $2$ epochs.}
\label{tab:second_epoch_business}
\scalebox{0.86}{
\begin{tabular}{c|c|c}
\hline
           & Short-Video Order & Short-Video Search LTV   \\
\hline
    Single-Epoch & 0.8489 & 0.8184 \\  \hline
  MEDA-NC & \textbf{0.8522} & \textbf{0.8248}  \\  \hline
  Improv. & +0.33\% & +0.64\%   \\ \hline
  MEDA-C & \textbf{0.8513} & \textbf{0.8233}  \\  \hline
  Improv. & +0.24\% & +0.49\%  \\ \hline
\end{tabular}
}
 
\end{table}

\begin{figure}[t!]
    \centering
    \includegraphics[width=\linewidth]{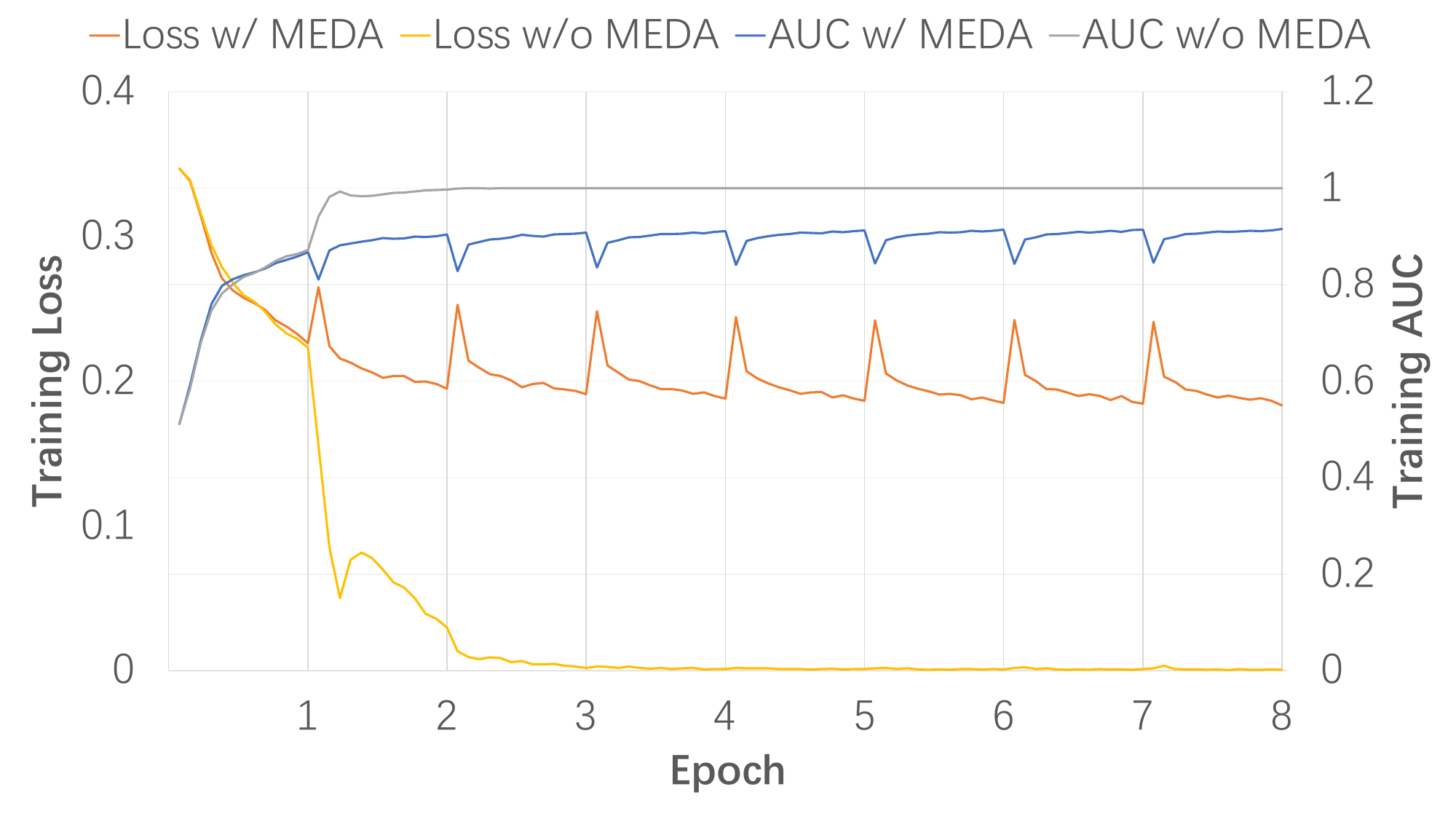}
    \caption{The training metric curves of training DNN on the Taobao dataset, with or without non-continual MEDA. }
    \label{fig:da_training}
\end{figure}

\begin{figure}[t!]
    \centering
    \includegraphics[width=\linewidth]{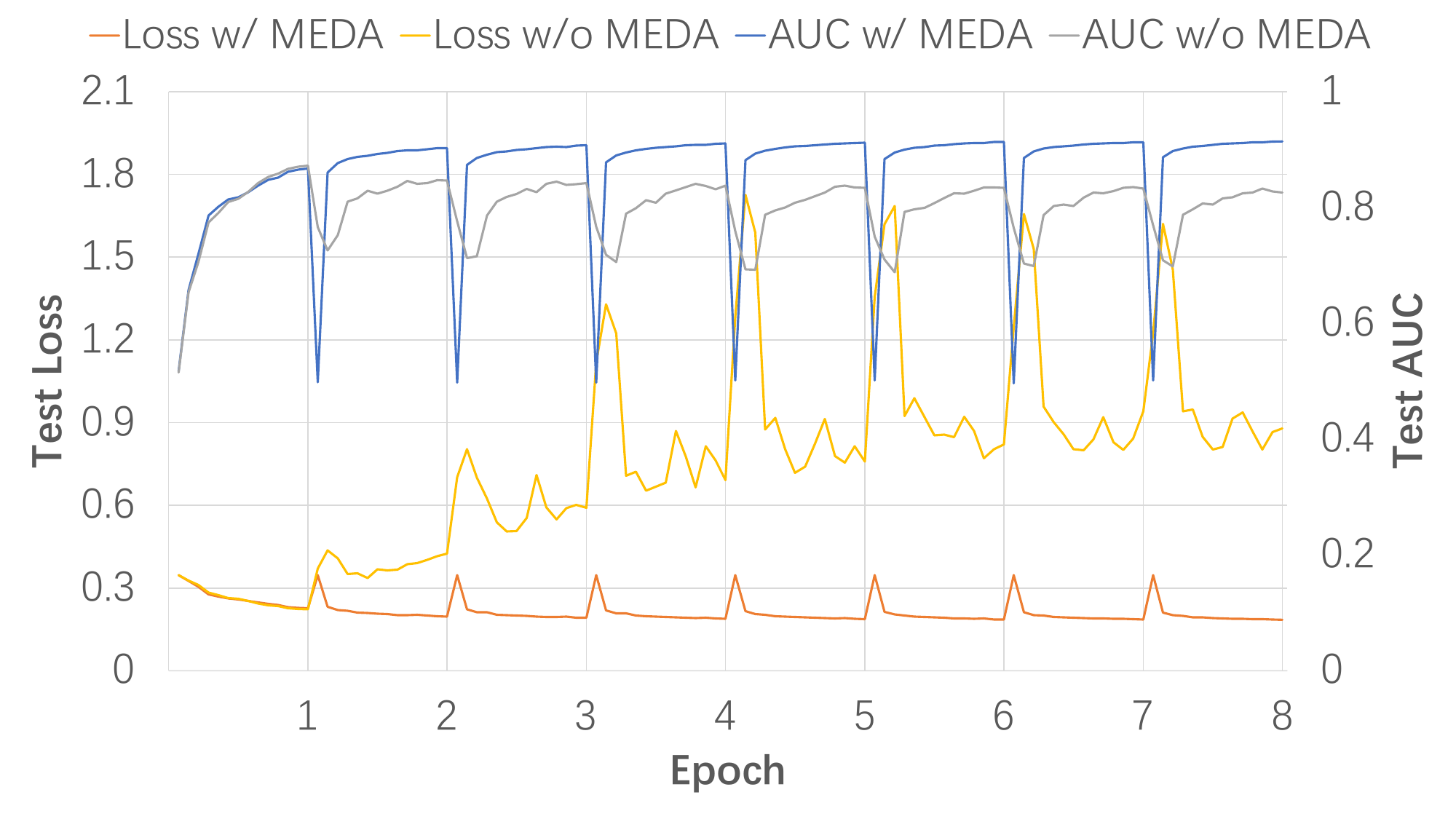}
    \caption{The test metric curves of training DNN on the Taobao dataset, with or without non-continual MEDA.  }
    \label{fig:da_test}
\end{figure}

\noindent\textbf{\emph{\underline{Effectiveness of Data Augmentation}}}.
Figures~\ref{fig:da_training} and~\ref{fig:da_test} illustrate the behavior of MEDA during training, where the use of MEDA results in a gradual decrease in training loss from the second epoch onwards, similar to encountering new data. At the beginning of each epoch, there is a slight increase in training loss, resembling the performance when facing new data from a different domain, leading to recognition errors. Furthermore, Tables~\ref{tab:how_many_epochs} and~\ref{tab:how_many_epochs_amazon} demonstrate that MEDA achieves comparable test AUC to single-epoch learning with fewer data. In most cases, MEDA with half the data surpasses single-epoch training with complete data, especially on Taobao, which validates the efficacy of data augmentation. The relatively better performance on Taobao compared to Amazon suggests that Amazon exhibits more severe data sparsity, resulting in weaker performance when joining multiple samples. Notably, in the case of ADFM on Taobao, even $3$ epochs with $1/8$ of the data outperform a single epoch with complete data. This may be attributed to ADFM's extensive behavior window and increased interactions between ID features, as MEDA enhances the importance of ID relationships, providing more opportunities for improvements.

\begin{table}[t!]
\centering
\caption{The numbers of training epochs required for non-continual MEDA on the Taobao dataset with different data keeping rates of $\rho$s to achieve the test AUC with one-epoch on the complete data and the corresponding test AUCs.}
\label{tab:how_many_epochs}
\scalebox{0.95}{
\begin{tabular}{c|c|c|c|c|c|c}
\hline
    $\rho$      &  & DNN & DIN & DIEN & MIMN & ADFM \\
\hline
\multirow{2}{*}{100\%} & \#Epochs & 1 & 1 & 1 & 1 & 1 \\ \cline{2-7}
 & Test AUC & 0.8714 & 0.8804 & 0.9032 & 0.9392 & 0.9462 \\ \hline
\multirow{2}{*}{50\%} & \#Epochs & \textbf{2} & \textbf{2} & \textbf{3} & \textbf{3} & \textbf{2} \\ \cline{2-7}
 & Test AUC & 0.8864 & 0.8989 & 0.9139 & 0.9444 & 0.9525 \\ \hline
\multirow{2}{*}{25\%} & \#Epochs & \textbf{4} & \textbf{3} & \textbf{6} & \textbf{13} & \textbf{2} \\ \cline{2-7}
 & Test AUC & 0.8802 & 0.8847 & 0.9048 & 0.9395 & 0.9466 \\ \hline
\multirow{2}{*}{12.5\%} & \#Epochs & \textbf{7} & \textbf{7} & \textbf{16} & $16^*$ & \textbf{3} \\ \cline{2-7}
 & Test AUC & 0.8716 & 0.8844 & 0.9030 & 0.9287 & 0.9470 \\ \hline
\end{tabular}}
\begin{tablenotes}
\footnotesize
\item[1] $^*$ means with the corresponding epoch number MEDA does not achieve the performance of single-epoch training on the complete data. 
\end{tablenotes}
\end{table}

\begin{table}[t!]
\centering
\caption{The same experiment as that conducted in Table~\ref{tab:how_many_epochs} on the Amazon dataset.}
\label{tab:how_many_epochs_amazon}
\scalebox{0.95}{
\begin{tabular}{c|c|c|c|c|c|c}
\hline
  $\rho$      &  & DNN & DIN & DIEN & MIMN & ADFM \\
\hline
\multirow{2}{*}{100\%} & \#Epochs & 1 & 1 & 1 & 1 & 1 \\ \cline{2-7}
 & Test AUC &0.8355 & 0.8477 & 0.8529 & 0.8686 & 0.8428 \\ \hline
\multirow{2}{*}{50\%} & \#Epochs & \textbf{10} & \textbf{4} & $16^*$ & \textbf{4} & \textbf{7} \\ \cline{2-7}
 & Test AUC & 0.8370 & 0.8551 & 0.8481 & 0.8879 & 0.8441 \\ \hline
\multirow{2}{*}{25\%} & \#Epochs & $16^*$ & $16^*$ & $16^*$ & $16^*$ & $16^*$ \\ \cline{2-7}
 & Test AUC & 0.8268 & 0.8446 & 0.8337 & 0.8578 & 0.8319 \\ \hline
\multirow{2}{*}{12.5\%} & \#Epochs & $16^*$ & $16^*$ & $16^*$ & $16^*$ & $16^*$ \\ \cline{2-7}
 & Test AUC & 0.8093 & 0.8262 & 0.8157 & 0.8328 & 0.8202 \\ \hline
\end{tabular}}
\end{table}

\begin{figure}[t!]
\centering
\subfigure[Cosine Between MLPs]{\includegraphics[width=0.23\textwidth]{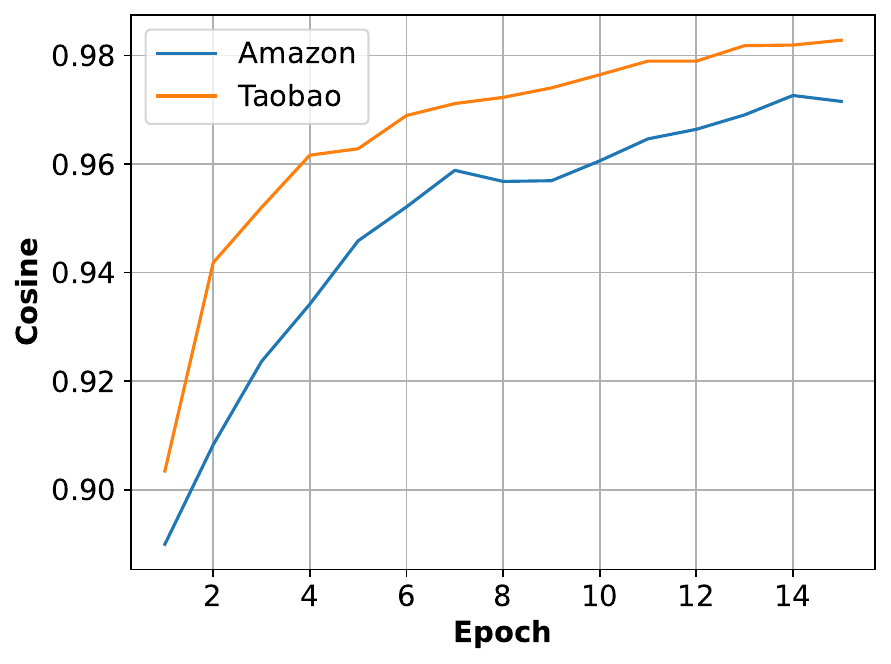}}
\subfigure[$\ell_2$ Distance Between MLPs]{\includegraphics[width=0.23\textwidth]{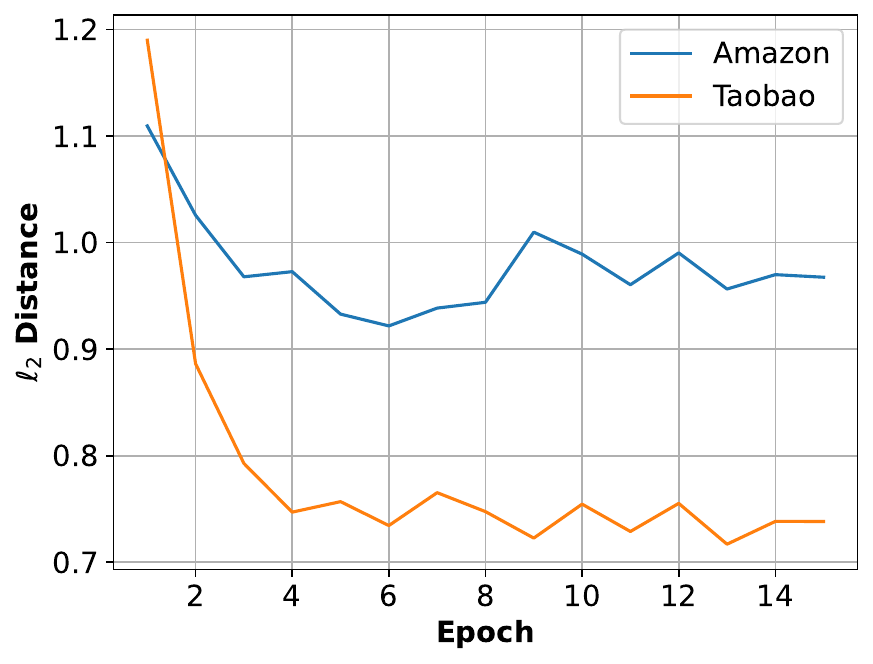}}
\subfigure[Cosine Between Embeddings]{\includegraphics[width=0.23\textwidth]{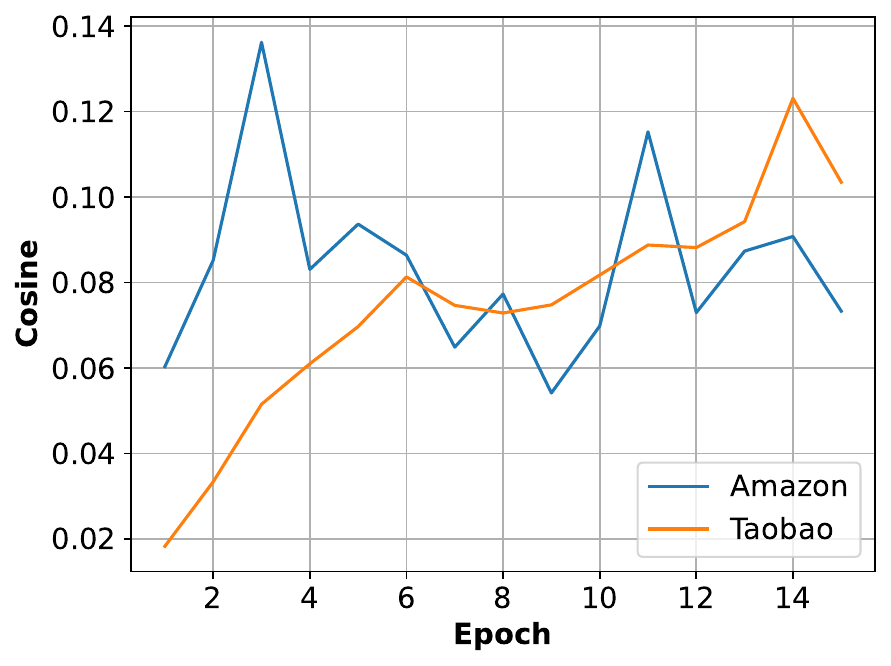}}
\subfigure[$\ell_2$ Distance Between Embeddings]{\includegraphics[width=0.23\textwidth]{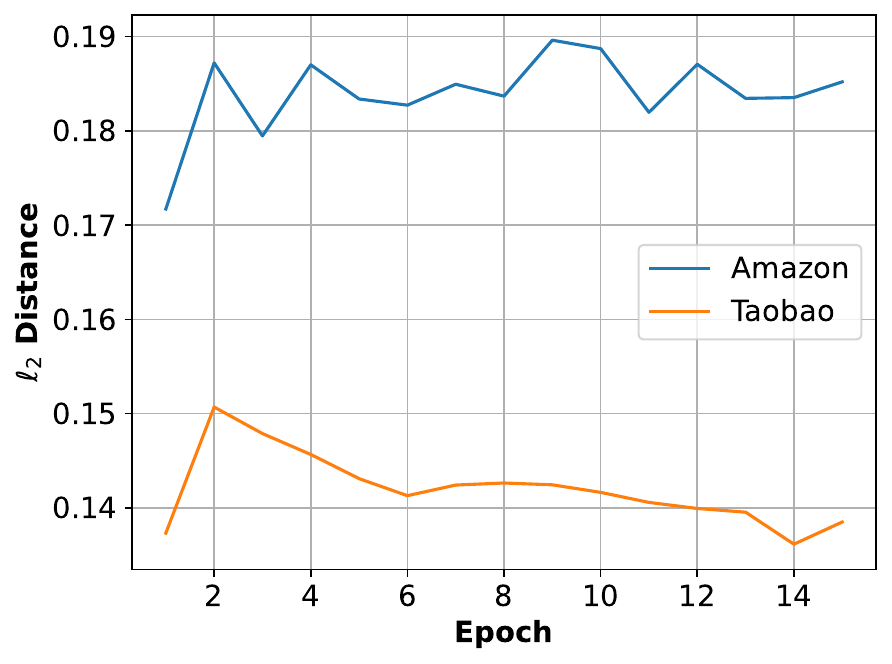}}
\caption{The parameter-convergence metric curves of non-continual MEDA using DNN on the public datasets.}
\label{fig:param_converge}
\end{figure}

\noindent\textbf{\emph{\underline{Effectiveness of MLP Convergence}}}. 
In Figure~\ref{fig:param_converge} (a) and (b), our MEDA approach is shown to enhance MLP convergence, as evidenced by the increasing similarity between two sets of MLP parameters in successive epochs. The cosine similarity between parameter groups continues to rise with each epoch, while the $\ell_2$ distance ceases to decrease after epoch 6, indicating that parameter direction is more crucial for CTR models than parameter distance. Moreover, the substantial discrepancy in final embedding parameters even at epoch 15 underscores the data augmentation effect of MEDA. Interestingly, the embedding parameters exhibit convergence on Taobao based on cosine similarity but not notably on Amazon, possibly due to the challenges posed by severe data sparsity in learning similar embedding patterns across epochs.

\subsection{Ablation Study}
\label{section:ablation}
Due to the space limit, some ablation studies are conducted on the Amazon dataset only, for whose overfitting issue is more severe.

\begin{figure}[t!]
    \centering
    \includegraphics[width=0.9\linewidth]{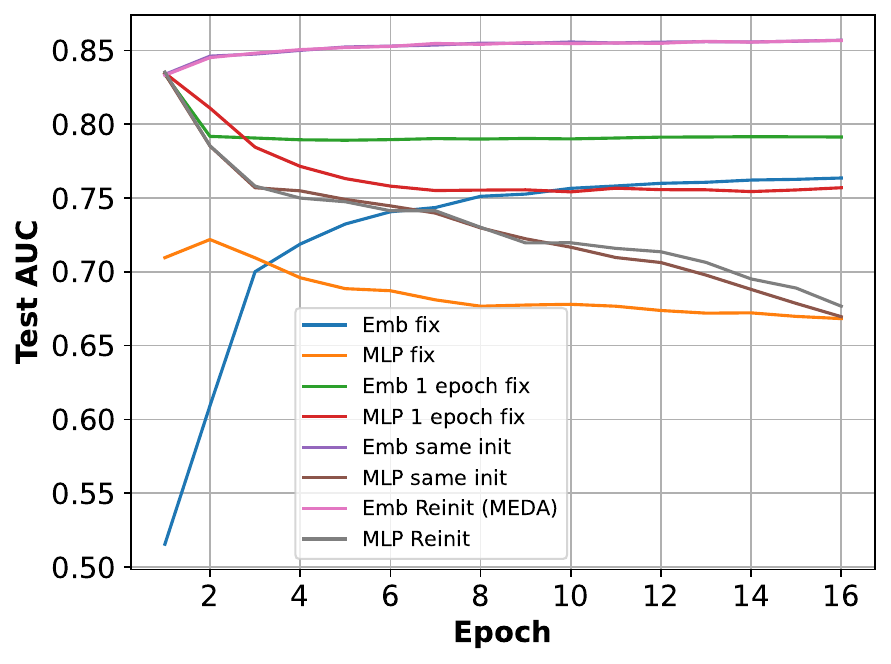}
    \caption{The test AUC curves for training DNN on the Amazon dataset with different training paradigms. }
    \label{fig:amazon_ablation}
\end{figure}

\noindent\textbf{\emph{\underline{Primary Responsibility of Embedding}}}. 
In Figure~\ref{fig:amazon_ablation}, we show that the primary causative factor of one-epoch overfitting is the embedding overfitting, where we compare $4$ groups of variants:

\begin{enumerate}
    \item Emb (MLP) fix: fixing the embedding (MLP) parameter and training the MLP (embedding).
    \item Emb (MLP) 1 epoch fix: after epoch $1$, fixing the embedding (MLP) and training the MLP (embedding); 
    \item Emb (MLP) same init: for each epoch, using the same initialization result to reinitialize the embedding (MLP);
    \item Emb (MLP) Reinit: for each epoch, independently reinitialize the embedding (MLP).
\end{enumerate}


The group ``Emb (MLP) fix'' reveals that the MLP does not overfit when the embedding is fixed, even containing considerable noise due to being solely initialized. Across all variants, severe overfitting occurs unless the embedding is controlled. When the embedding is controlled, overfitting can be mitigated (group 2) or avoided altogether (groups 1, 3, and 4). These findings underscore that one-epoch overfitting primarily stems from embedding overfitting.

\begin{figure}[t!]
    \centering
    \includegraphics[width=0.62\linewidth]{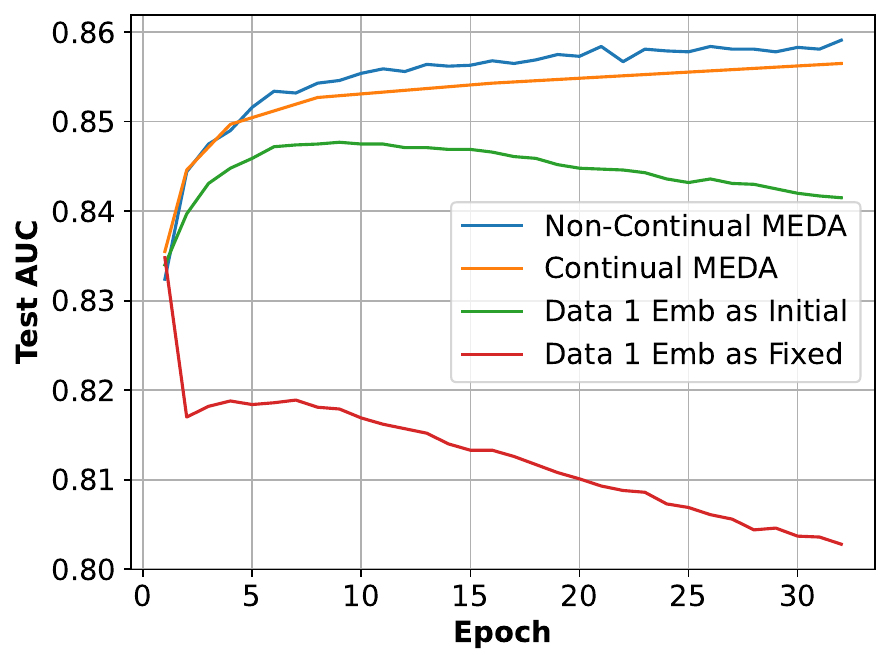}
    \caption{The test AUC curves of DNN on the Amazon dataset, comparing different variants of continual MEDA to train $\mathcal{D}_{tr}^2$ multiple times. The continual MEDA has run 2, 4, 8, 16, and 32 epochs. Note that the results of MEDA methods are only for reference because they also train $\mathcal{D}_{tr}^1$ multiple times.  }
    \label{fig:continual_ablation}
\end{figure}

\noindent\textbf{\emph{\underline{Variants of Continual MEDA}}}. 
In Figure~\ref{fig:continual_ablation}, we test two variants of our Continual MEDA:
\begin{enumerate}
    \item Data 1 Emb as Initial: using the final embedding of training $\mathcal{D}_{tr}^1$ as the initial embedding for multi-epoch training of $\mathcal{D}_{tr}^2$.
    \item Data 1 Emb as Fixed: after training $\mathcal{D}_{tr}^2$, using and fixing the final embedding of training $\mathcal{D}_{tr}^1$ for multi-epoch training of $\mathcal{D}_{tr}^2$.
\end{enumerate}


Both variants exhibit overfitting beyond epoch 10, indicating that when the MLP is trained repeatedly with the same embedding trained on data, the dependency between them intensifies, resulting in overfitting. This highlights the significance of our continual MEDA approach in reducing the dependence between the embedding and the MLP.
 
We also test two variants of our Continual MEDA on four datasets in Table~\ref{tab:second_epoch_ablation}:

\begin{enumerate}
    \item MEDA-C Emb-Reuse: same with the ``Data 1 Emb as Initial''.
    \item MEDA-C Multi-MLP: same with continual MEDA, except for that when training $\mathcal{D}_{tr}^1$ for $\E_1^2$, we reinitialize the MLP.  
\end{enumerate}

The results show that, compared with continual MEDA, both training $\mathcal{D}_{tr}^1$ only once and using another MLP's embedding compromise the performance. Nonetheless, both variants still outperform ``Single-Epoch'', showing the effectiveness of multi-epoch learning.

\begin{table}[t!]
\centering
\caption{The test AUC results on four datasets, comparing different variants of continual MEDA to train $\mathcal{D}_{tr}^2$ multiple times. Except for ``Single-Epoch'', other methods run $2$ epochs.}
\label{tab:second_epoch_ablation}
\scalebox{0.95}{
\begin{tabular}{c|c|c|c|c}
\hline
        & Amazon & Taobao  & SVO & SVSL   \\
\hline
    Single-Epoch &   0.8355 & 0.8714 & 0.8489 & 0.8184 \\  
  MEDA-NC &   {0.8450} & {0.9034} & 0.8522 & 0.8248  \\  \hline
  MEDA-C &   {0.8446} & {0.9054} & 0.8517 & 0.8233\\  
  MEDA-C Emb-Reuse & 0.8397  & {0.8943} & 0.8512 & 0.8221\\  
  MEDA-C Multi-MLP & 0.8382  & {0.8939} & 0.8502 & 0.8213\\   \hline
\end{tabular}
}
\end{table}

\begin{figure}[t!]
    \centering
    \includegraphics[width=0.62\linewidth]{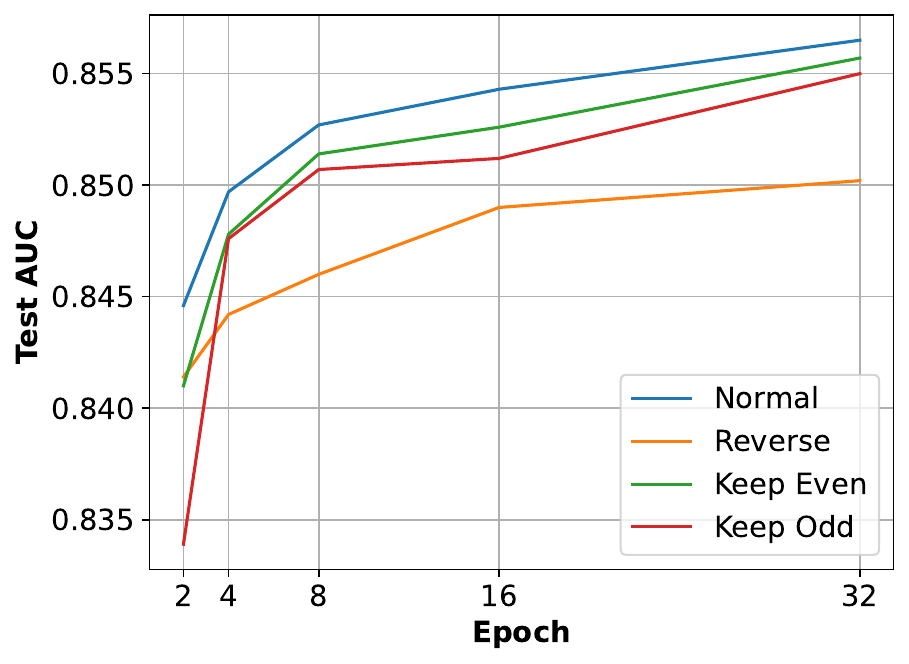}
    \caption{The test AUC curves with DNN on the Amazon dataset, comparing different variants of our continual MEDA. }
    \label{fig:continual_order}
\end{figure}

\noindent\textbf{\emph{\underline{Hyperparameter Robustness}}}. 
In Figure~\ref{fig:continual_order}, we examine the impact of reversing the embedding order or omitting some (even or odd) groups of embeddings in training 
$\mathcal{D}_{tr}^2$
  using our continual MEDA. Both types of variants result in compromised performance, indicating that either training on old embeddings or reducing data augmentation is inferior. Nonetheless, these variants still demonstrate enhanced performance and robustness of our MEDA methodology across increasing epochs without overfitting.

\subsection{Online Result}
\label{section:online_ab}
We conduct an online A/B test on Kuaishou's industrial video advertising platform, focusing on retention prediction. This platform processes billions of user requests daily, with millions of item candidates. It incorporates a four-stage recommender system: candidate retrieval, pre-ranking, ranking, and reranking, each progressively narrowing down the item selection for users. We deploy MEDA in the ranking module. The baseline method performs single-epoch learning, while we adopt our continual MEDA to conduct the online A/B experiment.
The experiment spanned 9 days, with 10\% of the total online traffic allocated for both the baseline and our MEDA approach. Results in Table~\ref{tab:abtest} demonstrate that MEDA significantly enhances the test AUC, user retention, and \emph{overall} platform rewards (as evaluated by revenue and revenue expected by clients). This marks the first solution addressing overfitting in multi-epoch training of large-scale sparse models for advertising recommendations. In our scenario, achieving satisfactory model performance typically necessitates training on one month's worth of data. Remarkably, with MEDA, comparable results can be obtained after just two weeks of training. Thus, implementing MEDA substantially reduces the required sample size and training costs while maintaining equivalent outcomes.


\begin{table}[t!]
\centering
\caption{Online A/B Test Performance.}
\label{tab:abtest}
\scalebox{0.86}{
\begin{tabular}{c|c|c|c|c}
\hline
           & Test AUC & Retention & Revenue & Expected Revenue \\
\hline
   Improv. & +0.14\% & +6.6\% & +0.32\% &  +0.91\% \\  \hline
\end{tabular}
}
 
\end{table}


\section{Discussion}
One limitation of this study is the use of independent random embedding initializations for multiple epochs, potentially discarding valuable information in the final embedding parameters from the first epoch. For instance, certain high-frequency categorical features like item category IDs may not be at risk of overfitting, suggesting that their corresponding embedding parameters should be retained. However, a potential solution exists to address this issue. Upon closer examination of our MEDA framework, a key constraint is that initial parameters must not contain \emph{exact} information from any data sample in upcoming training data. This constraint aligns with principles of differential privacy~\citep{chaudhuri2011differentially,dwork2014algorithmic}, where the goal is to limit the inference of \emph{exact} information from trained parameters. By treating the random initialization process as noise addition, tailored noise levels can be designed based on the data frequency of each categorical feature. This approach could enhance MEDA's performance further, offering an avenue for future exploration.

Conventional wisdom might suggest that multi-epoch learning could sacrifice timeliness in pure online learning~\cite{burlutskiy2016investigation}. Yet, it is a shared problem of all multi-epoch learning approaches and is mainly due to limited computation resources. Recent industrial advances combine batch and online learning to achieve both accuracy and timeliness, then multi-epoch learning including MEDA can contribute to batch learning to meet resource constraints.

\section{Conclusion}
In this paper, we propose a novel and simple Multi-Epoch learning with
Data Augmentation (MEDA) framework, covering both non-continual and continual learning settings.
The experimental results of both public and business datasets show that MEDA effectively achieves the desired effect of data augmentation and multi-epoch learning can outperform
the conventional single-epoch training by a significant margin.
Furthermore, MEDA's deployment in a real-world online advertising system and subsequent A/B testing demonstrate its substantial benefits in practical applications.
\balance
\bibliographystyle{ACM-Reference-Format}
\bibliography{sample-base}

\end{document}